# A Cross-institutional Evaluation on Breast Cancer Phenotyping NLP Algorithms on Electronic Health Records

Sicheng Zhou, MS, Institute for Health Informatics, University of Minnesota, Minneapolis, MN, USA

Nan Wang, School of Statistics, University of Minnesota, Minneapolis, MN, USA

Liwei Wang, MD, PhD, Department of AI and Informatics Research, Mayo Clinic, Rochester, MN, USA

Ju Sun, PhD, Department of Computer Science & Engineering, University of Minnesota, Minneapolis, MN, USA

Anne Blaes, Department of Medicine, University of Minnesota, Minneapolis, MN, USA

Hongfang Liu, PhD, Department of AI and Informatics Research, Mayo Clinic, Rochester, MN, USA

Rui Zhang, PhD, Division of Computational Health Sciences, Department of Surgery, University of Minnesota, Minneapolis, MN, USA

**Abstract**

**Objective**: The generalizability of clinical large language models is usually ignored during the model development process. This study evaluated the generalizability of BERT-based clinical NLP models across different clinical settings through a breast cancer phenotype extraction task.

**Materials and Methods**: Two clinical corpora of breast cancer patients were collected from the electronic health records from the University of Minnesota and the Mayo Clinic, and annotated following the same guideline. We developed three types of NLP models (i.e., conditional random field, bi-directional long short-term memory and CancerBERT) to extract cancer phenotypes from clinical texts. The models were evaluated for their generalizability on different test sets with different learning strategies (model transfer vs. locally trained). The entity coverage score was assessed with their association with the model performances.

**Results**: We manually annotated 200 and 161 clinical documents at UMN and MC, respectively. The corpora of the two institutes were found to have higher similarity between the target entities than the overall corpora. The CancerBERT models obtained the best performances among the independent test sets from two clinical institutes and the permutation test set. The CancerBERT model developed in one institute and further fine-tuned in another institute achieved reasonable performance compared to the model developed on local data (micro-F1: 0.925 vs 0.932).

**Conclusions**: The results indicate the CancerBERT model has the best learning ability and generalizability among the three types of clinical NLP models. The generalizability of the models was found to be correlated with the similarity of the target entities between the corpora.

**INTRODUCTION**

With the high adoption rate of electronic health records (EHR), natural language processing (NLP) methods have been increasingly developed to leverage clinical texts for clinical decision support and clinical research [1]. They were widely adopted in applications such as real-time cancer case identification [2], classify medical prescriptions [3], and automatic extraction of heart failure from EHR [4]. The developed NLP methods can be mainly classified into either symbolic or statistical approaches [5]. The symbolic approaches dominated the clinical domain in the early years since they could meet the basic information needs of many applications in the clinical domain while not needing a large amount of annotated data, which requires intensive human labor [6]. Meanwhile, their results are easy to interpret [6]. One major drawback of symbolic approaches is the lack of portability [1]. In recent years, benefiting from the development of large language models (transformer-based models [7]) and increased annotation data in the clinical domain, the statistical approaches were developed rapidly and achieved impressive performances for different clinical NLP tasks. However, there's a huge gap between model performances and the understanding of the generalizability of these large language models [8]. The successfully deployed clinical NLP systems are often developed based on data from a single healthcare institute, their performances can be various if applied in different healthcare institutes [9]. The EHR platforms and clinical documentation rules and conventions vary among different healthcare institutes, which may result in different clinical texts even when documenting highly similar medical events. The clinical language variations contain both syntactic variations and semantic variations [10], and these variations could accumulate in the whole clinical corpus. The influence of these variations among different clinical corpora on the generalizability of NLP models developed in a single healthcare institute remains an important research question.

Few studies have explored the generalizability of clinical NLP models. For example, Sohn et al. evaluated the portability of a rule-based NLP system that identifies the asthma patient from the clinical cohort [1]. The NLP system was developed based on a cohort from a single hospital was evaluated using an external cohort and the performance of the NLP system dropped significantly due to the clinical documentation variations [1]. Another study developed a rule-based NLP system to identify patients with a family history of pancreatic cancer and tested it on data from two clinical institutions [11]. The study found that for very specific NLP tasks, the rule-based system is portable as long as the rules are kept simple and could be updated using the new data. Liu et al [12] evaluated the performances of the NLP system to detect the smoking status across different institutes. The system consists of both machine learning and rule-based modules. It was found that moderate efforts were needed to make the NLP system portable, the efforts include annotating more data to further train the machine learning module, and adding new rules based on the new data. These studies mainly focused on evaluating the generalizability of rule-based and traditional machine learning methods, currently, there's a lack of study that explore the generalizability of transformer-based models in the clinical domain [8].

The transformer-based models have achieved excellent performances for different clinical NLP tasks, however, the development of these models needs a large amount of computing resources and human labor for data annotation. Also, the annotated clinical data cannot be shared in most cases due to privacy issues, which set obstacles to developing generalizable models. Understanding the generalizability of Transformer based models is important in clinical NLP domain, lots of labor and computational resources could be saved by avoiding training similar models if the models are generalizable and portable among different clinical institutes. Recently, we have developed a cancer specific language model, i.e., CancerBERT, that aims to identify eight breast cancer phenotypes using clinical texts of breast cancer patients collected from the University of Minnesota's M Health Fairview (UMN) [13]. The CancerBERT was developed based on the Bidirectional Encoder Representations from Transformers (BERT) language model [14], and achieved excellent performances on the breast cancer phenotypes extraction task. However, the CancerBERT models were developed and evaluated solely on the EHR corpus from UMN. The objective of this study is to explore the generalizability of CancerBERT models by evaluating their performances on the corpus collected from another institution, the Mayo Clinic (MC). We also evaluated and compared with other benchmark models, including the conditional random field models (CRF) and Bi-directional long-short memory CRF (BiLSTM-CRF) models. We evaluated the generalizability of the models through a clinical information extraction task. The generalizability of the models was identified to be correlated with the similarities of target concepts between the corpora. The contributions of this study include:

1. We evaluated the impact of corpus heterogeneity on generalizability of BERT-based NLP models.

2. We also constructed a permutation dataset to analyze the robustness of the models.

3. We compared two strategies for transferring models between clinical institutes, i.e., i) direct transfer vs ii) continuous fine-tuning.

**METHODS AND MATERIALS**

**Overview of the study**

This study was approved by the institutional review boards (IRB) of UMN and MC. **Figure 1** shows the pipeline of the study. The clinical texts of breast cancer were collected from EHR in the two institutes (UMN and MC). MC team annotated their corpus by following the same annotation guideline as previously defined by the UMN team [13]. The CancerBERT$_{UMN}$ models, along with the benchmark CRF$_{UMN}$ and BiLSTM$_{UMN}$ models developed on the UMN corpus were originally designed to extract the eight types of breast cancer phenotypes (i.e., *Hormone receptor type*, *Hormone receptor status*, *Tumor size*, *Tumor site*, *Cancer grade*, *Histological type*, *Cancer laterality* and *Cancer stage)* from the clinical texts. We externally evaluated the performances of models trained on UMN site using data from the MC site. In addition, we compared two transfer learning strategies for the NLP models: (1) continuously trained (MC revised models) and (2) solely locally-trained (MC model) on the MC site. In addition, we conducted two additional experiments to explore the advantages of BERT-based models compared to the traditional BiLSTM-CRF and CRF models as baseline models in terms of model robustness.

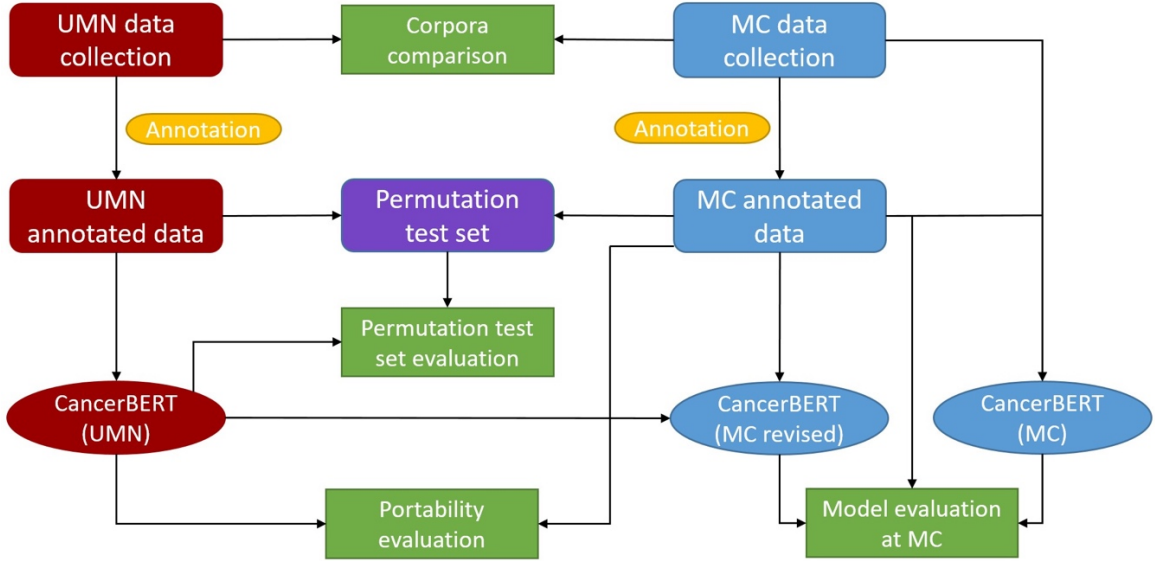

**Figure 1**. Pipeline of the study. Data were collected and annotated from the UMN and MC. The UMN models were externally evaluated on the MC data. UMN models were further revised on MC data and evaluated as comparison.

**Data Sources**

The data used in this study was obtained from the EHRs of two clinical institutes, i.e., the UMN Clinical Data Repository, and the MC. The EHR data from the UMN contains the health records of 17,970 breast cancer patients from the year 2001 to 2018. The EHR data from the MC contain 54,050 breast cancer patients from 2000 to 2020.

**Manual Annotation and corpora comparison**

The annotation schema was introduced in the previous study [13], we used the same schema to annotate the clinical texts extracted from MC EHR. We randomly sampled one hundred breast cancer patients, and for each patient, we randomly sampled one clinical note and one pathology report within 90 days after diagnosis date for annotation. This resulted in 81 clinical notes and 80 pathology reports. Two annotators worked on 10% of the text annotation, the inter-annotator agreement (IAA) was evaluated as Cohen's Kappa of 0.80. Then, one annotator completed all annotation tasks.

To better understand the impact of corpus across sites on the performances of NLP methods. We compared the variations of the annotated clinical texts from two sites. The basic corpus statistics such as the number of sentences, the average length of the sentences and the unique number of tokens were summarized and compared. Also, we measured the similarity between the two corpora by representing the two corpora as the normalized term frequency inverse document frequency (TF-IDF) vectors and calculating the cosine similarity [1]. Each corpus vector contains the average TF-IDF values for unique terms in the corpus. The average TF-IDF value of term t is defined as equation 1, where N is the total number of documents in corpus, the $TF_i(t)$ is the frequency of token $t$ in $i$th document divided by the total number of tokens in the *i*th document, and the *IDF(t)* is the *log(N)* divided by the number of documents contains token *t*.

$$TF-IDF(t) = \sum_{i=1}^{N} TF_i(t) * \frac{IDF(t)}{N} \qquad (1)$$

Besides the corpus level similarity, we also calculated the similarity of each breast cancer phenotype category (phenotype similarity) between the two corpora. Similarly, each breast cancer phenotype was represented by a normalized TF-IDF vector that contains the average TF-IDF values of unique terms under each breast cancer phenotype category. Cosine similarity was calculated for the same breast cancer phenotype category from the two corpora.

**Development of NLP models for breast cancer phenotypes extraction task**

We evaluated the generalizability of three types of machine learning models in this study, i.e., CRF, BiLSTM-CRF and CancerBERT [13] models. The CRF and BiLSTM-CRF are conventional machine learning models that have been proven effective for the name entity recognition (NER) tasks in the clinical domain [15-17]. For CRF and BiLSTM baseline models, we experimented with both GloVe (trained on Wikipedia) [18] and Word2vec (trained on Google news) embeddings [19] as input. The CancerBERT models were developed based on the BERT language model and showed superior performance compared to the CRF and BiLSTM-CRF models for our breast cancer phenotype extraction task [13].

**Portability evaluation for breast cancer phenotype extraction models**

MC annotated data were divided into a training set (60%), development set (10%), and test set (30%). In this study, we evaluated the performance of the following three sets of models on the MC corpus test set.

1. Models originally trained only on UMN corpus [13]. These UMN models ($CRF_{UMN}$, $BiLSTM\text{-}CRF_{UMN}$, $CancerBERT_{UMN\_397}$, $CancerBERT_{UMN\_997}$) were evaluated on the MC data (test set) to test their portability through the task of extracting the breast cancer phenotypes from the clinical texts. All other models (e.g., $CancerBERT_{MC\_refined}$ and $CancerBERT_{MC\_397}$ introduced in the previous section) were also tested as a comparison.

2. Continuously fine-tune UMN models on MC corpus, including $CRF_{MC\_refined}$, $BiLSTM\text{-}CRF_{MC\_refined}$, and $CancerBERT_{MC\_refined}$. The $CancerBERT_{UMN\_397}$ model was further fine-tuned on MC annotated data to obtain the MC refined $CancerBERT_{MC\_refined}$ model. To refine the UMN models, all the UMN locally trained models were fine-tuned on the training and development sets to obtain the corresponding MC refined models.

3. CancerBERT model trained only on the MC corpus (without UMN data). Following the same steps [13], we developed another $CancerBERT_{MC\_397}$, which was pre-trained only on the MC corpus (contains about 5 million clinical documents) and fine-tuned on the training and development sets of the MC data as a benchmark model.

The breast cancer phenotype extraction can be framed as a NER task, and the phenotype level (name entity) evaluation was applied. We used the micro-average (assign equal weight to each sample) F1 and macro-average (assign equal weight to each category) F1 as metrics, and both exact match and lenient match were used following the i2b2standard [20]. F1 score was calculated as the harmonic mean of the precision and recall scores, and provides a better assessment of model performance when data is imbalanced. All experiments were conducted 10 times, and the one-way ANOVA tests were conducted to compare the performances among different models.

**Permutation test set evaluation for UMN models**

The permutation dataset was used to simulate variations in the data and provides a better sense of generalizability of models when encounter new data [8, 21]. Thus, to explore how the data variances influence the performances of the models, we manually created the permutation test set, where we evaluated the BiLSTM-CRF$_{UMN}$, CRF$_{UMN}$ and CancerBERT$_{UMN}$ models. The entity permutation was used since entities are the core of NER task. To create the permutation test set, we first found all unique entities in MC annotated data that are not in the UMN annotated data; then we replaced the entities in the UMN test set with entities that were randomly sampled from the MC unique entities under the same category. In such a way, the permutation dataset contains new combinations of target entities and relevant contexts. Evaluation of models on the permutation test set could reveal their abilities to learn the contextual information of entities rather than memorize the entities. All models developed using the UMN data were evaluated using the permutation test set. The evaluation schema is the same as the portability evaluation of the models introduced in the previous section.

**Evaluation of model generalizability with entity coverage ratio (ECR)**

The portability of machine learning models is usually evaluated through measuring the change of standard metrics (e.g., F1, precision, recall) when tested on different test sets. However, this approach can only provide high-level insights, it fails to analyze the portability in a fine-grained way [22]. To analyze the model portability in a different perspective, we adapted the ECR proposed in previous study for our NER task [22]. The ECR measures to what degree the target entity in the test set has been seen in the training set with the same category. It was defined as equations (2) and (3):

$$ECR(e_i) = \begin{cases} 0, & C = 0 \\ \sum_{k=1}^{K} \frac{\#(e_i^{tr,k})\#(e_i^{te,k})}{C^{tr}}/C^{te}, & otherwise \end{cases} \quad (2)$$

Where $C^{tr} = \sum_{k=1}^{K} \#(e_i^{tr,k})$ and $C^{te} = \sum_{k=1}^{K} \#(e_i^{te,k})$. $e_i$ refers to a test entity, $\#(e_i^{tr,k})$ is the number of entity $e_i$ in the training set with label $k$, $\#(e_i^{te,k})$ is the number of entity $e_i$ in the test set with label $k$. The ECR scores are ranging from 0 to 1, indicates the difficulty to predict an entity (from easy to difficult) [22].

In our study, we calculated the ECR scores for the entities in the test set of UMN annotated data. We further divided the entities in the test set into different groups based on their ECR values, i.e., 0<= ECR <0.33, 0.33<= ECR <0.67, 0.67 <= ECR<1 and ECR = 1. We evaluated the UMN models (i.e., CancerBERT$_{UMN\_397}$, BiLSTM-CRF$_{UMN}$ and CRF$_{UMN}$ models developed solely on UMN data) on the UMN test set and investigated the relationships between ECR values and the model's performance on corresponding entities.

**Results**

**Comparison of UMN vs MC corpora**

The corpora of UMN and MC were compared from various perspectives (**Table 1**). Overall, the UMN corpus has more clinical documents per patient (190.5 vs 140.8), and each clinical document is shorter than MC on average (259 tokens vs 361 tokens). The number of breast cancer phenotypes annotated from the clinical texts of two sites. The numbers in the parenthesis are the unique number of terms in each phenotype entity. More entities related to breast cancer phenotypes were annotated from the UMN data. The IAA scores for annotations are 0.91 and 0.80 at UMN and MC sites, respectively. The similarities of phenotypes between two corpora were also calculated using the TF-IDF. Overall, the breast cancer phenotypes have higher similarity score compared to the overall corpus (phenotype average: 0.9088 vs overall: 0.5411).

*Table 1– Statistics for annotated pathology reports and clinical texts from UMN and MC. Numbers in parenthesis indicate the unique number of terms for each phenotype concept. Phenotype Similarity refers to the cosine similarity of each breast cancer phenotype category between corpora.*

|  | UMN | MC | Phenotype Similarity |
|---|---|---|---|
| Number of annotated documents | 200 | 200 | NA |
| Number of sentences | 10452 | 6165 | NA |
| Number of tokens | 266079 | 110,980 | NA |
| Number of unique tokens | 9151 | 5735 | NA |
| Average number of tokens in the sentence | 25.5 | 18.0 | NA |
| Hormone receptor type | 1673 (29) | 417 (18) | 0.9923 |
| Hormone receptor status | 436 (14) | 178 (12) | 0.9815 |
| Tumor size | 540 (305) | 393 (262) | 0.8076 |
| Tumor site | 329 (173) | 187 (135) | 0.7399 |
| Cancer grade | 271 (15) | 234 (23) | 0.8810 |
| Cancer laterality | 1192 (4) | 1846 (10) | 0.9833 |
| Cancer stage | 173 (38) | 207 (55) | 0.9965 |
| Histological type | 1070 (95) | 726 (77) | 0.8880 |

**Portability evaluations of machine learning-based models and BERT models**

The portability evaluation results (both strict match and lenient match F1 scores) for the two classic NER models, i.e., CRF and BiLSTM-CRF, are shown in **Table 2**. Two word embeddings were used as input (Glove Wikipedia 6B [18] and Word2Vec Google News [19]), the models were evaluated using the MC test set, and the results show the performances of models that directly obtained from the UMN site ($CRF_{UMN}$, $BiLSTM-CRF_{UMN}$) and also the models that were further fine-tuned on the MC data ($CRF_{MC\_refined}$, $BiLSTM-CRF_{MC\_refined}$).

*Table 2– Portability evaluation results (strict match (lenient match) F1 scores) for CRF and BiLSTM-CRF models. Note that models with subscript "UMN" indicate models trained only on UMN corpus and models with subscript "MC_refined" are UMN models with continuous fine tune on MC corpus.*

| Word embeddings | Glove Wikipedia 6B | | Glove Wikipedia 6B | | Word2Vec Google News | | Word2Vec Google News | |
|---|---|---|---|---|---|---|---|---|
| Models | $CRF_{UMN}$ | $CRF_{MC\_refined}$ | BiLSTM-CRF $_{UMN}$ | BiLSTM-CRF $_{MC\_refined}$ | $CRF_{UMN}$ | $CRF_{MC\_refined}$ | BiLSTM-CRF $_{UMN}$ | BiLSTM-CRF $_{MC\_refined}$ |
| Hormone Receptor type | 0.939 (0.941) | 0.944 **(0.954)** | 0.925 (0.925) | 0.943 **(0.954)** | 0.945 (0.948) | **0.948** (0.951) | 0.939 (0.939) | 0.918 (0.929) |
| Hormone Receptor status | 0.527 (0.527) | 0.542 (0.542) | 0.794 (0.794) | **0.876*** **(0.876*)** | 0.529 (0.529) | 0.491 (0.491) | 0.867 (0867) | 0.837 (0.837) |
| Tumor size | 0.224 (0.294) | 0.694 (0.749) | 0.392 (0.509) | **0.711*** **(0.813*)** | 0.244 (0.350) | 0.663 (0.724) | 0.367 (0.421) | 0.592 (0.738) |
| Tumor site | 0.053 (0.136) | 0.303 (0.600) | 0.266 (0.400) | 0.472 (0.758) | 0.044 (0.128) | 0.321 (0.596) | 0.205 (0.323) | **0.361*** **(0.668*)** |
| Tumor grade | 0.647 (0.681) | 0.881 **(0.925*)** | 0.903 (0.903) | **0.916*** (0.916) | 0.647 (0.685) | 0.861 (0.913) | 0.849 (0.849) | 0.881 (0.881) |
| Tumor laterality | 0.846 (0.846) | 0.935 (0.935) | 0.882 (0.882) | **0.952*** **(0.952*)** | 0.853 (0.853) | 0.934 (0.934) | 0.874 (0.874) | 0.930 (0.930) |
| Cancer stage | 0.773 (0.773) | 0.868 (0.868) | 0.578 (0.578) | **0.891*** **(0.891*)** | 0.632 (0.632) | 0.873 (0.873) | 0.593 (0.593) | 0.838 (0.838) |
| Histological type | 0.829 (0.896) | **0.937** (0.964) | 0.847 (0.917) | 0.934 (0.959) | 0.845 (0.907) | 0.934 **(0.965)** | 0.839 (0.921) | 0.899 (0.917) |
| F1 Macro average | 0.605 (0.637) | 0.763 (0.817) | 0.698 (0.738) | **0.837*** **(0.889*)** | 0.593 (0.629) | 0.753 (0.806) | 0.691 (0.723) | 0.782 (0.842) |
| F1 Micro average | 0.803 (0.822) | 0.883 (0.905) | 0.785 (0.814) | **0.893*** **(0.922*)** | 0.804 (0.825) | 0.880 (0.903) | 0.777 (0.802) | 0.853 (0.885) |

*Note: The scores were averaged scores after 10 runs, text in bold indicates highest performance, * indicates statistically higher than other methods (CI: 0.95).*

The portability evaluation results (strict match (lenient match) F1 scores) for the BERT-based models are shown in **Table 3**. The CancerBERT models developed at UMN (CancerBERT$_{UMN}$), along with two benchmark BERT-based models (BERT-large origin [23] and BlueBERT [14]) were evaluated. The CancerBERT$_{UMN}$ model has two variants, one with frequency-based 997 customized words (CancerBERT$_{UMN\_997}$) in its vocabulary and another one has 397 knowledge-based customized words (CancerBERT$_{UMN\_397}$). Each UMN model (**Table 3**, column *UMN*) was directly evaluated on the MC test set. Then these models were further fine-tuned on MC data to obtain MC refined models (**Table 3**, column *MC refined*) and evaluated again. In addition, the CancerBERT$_{MC\_397}$ model trained solely on the MC corpus was also evaluated for comparison, it gives a comprehensive overview for the performances of models developed using different transfer learning strategies.

*Table 3– Portability evaluation results (strict match (lenient match) F1 scores) for BERT-based models on MC test data set. Column of "UMN" includes models only trained on UMN corpus, and the column "MC_refined" contains models with fine tuen on MC corpus. The column of "MC only" is the model trained only on MC corpus.*

| Entity type | BERT-large Origin | | BlueBERT (PubMed+MIMIC III) | | CancerBERT UMN_997 | | CancerBERT UMN_397 | | CancerBERT MC_397 |
|---|---|---|---|---|---|---|---|---|---|
| | UMN | MC refined | UMN | MC refined | UMN | MC refined | UMN | MC refined | MC only |
| Hormone Receptor type | 0.926 (0.935) | 0.967 (0.969) | 0.897 (0.911) | 0.977 (0.977) | 0.923 (0.963) | 0.984 (0.988) | 0.942 (0.947) | 0.975 (0.981) | **0.993*** **(0.993*)** |
| Hormone Receptor status | 0.816 (0.816) | 0.910 (0.910) | 0.897 (0.897) | 0.932 (0.932) | 0.842 (0.842) | 0.901 (0.901) | 0.819 (0.819) | 0.926 (0.926) | **0.943*** **(0.943*)** |
| Tumor size | 0.633 (0.737) | 0.837 (0.903) | 0.440 (0.648) | 0.839 (0.915) | 0.595 (0.664) | 0.765 (0.813) | 0.745 (0.781) | **0.864** **(0.928*)** | 0.862 (0.907) |
| Tumor site | 0.666 (0.739) | 0.609 (0760) | 0.308 (0.671) | 0.590 (0.790) | 0.186 (0.742) | **0.733*** (0.792) | 0.709 (0.759) | 0.601 (0.786) | 0.661 **(0.832*)** |
| Tumor grade | 0.827 (0.827) | 0.909 (0.922) | 0.886 (0. 886) | 0.859 (0.939) | 0.869 (0.869) | 0.891 (0.891) | 0.846 (0.846) | **0.927*** **(0.943*)** | 0.863 (0.933) |
| Tumor laterality | 0.896 (0.896) | 0.928 (0.928) | 0.936 (0.936) | 0.954 (0.954) | 0.928 (0.928) | 0.939 (0.939) | 0.903 (0.903) | 0.959 (0.959) | **0.962** **(0.962)** |
| Cancer stage | 0.774 (0.774) | 0.934 (0.934) | 0.799 (0.799) | **0.953** **(0.953)** | 0.806 (0.806) | 0.870 (0.870) | 0.829 (0.829) | 0.949 (0.949) | 0.950 (0.950) |
| Histological type | 0.793 (0.888) | 0.926 (0.950) | 0.794 (0.902) | 0.931 (0.958) | 0.828 (0.923) | 0.849 (0.922) | 0.815 (0.914) | 0.934 (0.965) | **0.950*** **(0.981*)** |
| Macro average | 0.724 (0.829) | 0.874 (0.908) | 0.744 (0.831) | 0.879 (0.927) | 0.747 (0.842) | 0.867 (0.889) | 0.828 (0.849) | 0.892 (0.930) | **0.898*** **(0.932)** |
| Micro average | 0.843 (0.877) | 0.905 (0.922) | 0.817 (0.876) | 0.917 (0.943) | 0.829 (0.886) | 0.903 (0.925) | 0.864 (0.906) | 0.925 (0.947) | **0.932*** **(0.952*)** |

*Note: The scores were averaged scores based on 10 runs, text in bold indicates highest performance, \* indicates statistically higher than other methods (CI: 0.95).*

Based on **Table 2** and **Table 3**, all the models show different levels of performance drop compared to the models' performances evaluated on the original UMN test set. And after further fine-tuning the models on the MC data, the performances significantly increased.

**Permutation dataset evaluation**

For each type of model, we chose the one with the best performance to evaluate on the permutation dataset. **Table 4** shows the performances of the models, and how much performances (F1 scores) changed compared to their performance on the normal test set. The results show that the CancerBERT$_{UMN\_397}$ model has the smallest performance drop compared to other models, which indicates its robustness when dealing with new data from another institute.

*Table 4– Evaluation results for permutation dataset*

| Models Entities | CRF$_{UMN}$ | Changed F1 score | BiLSTM-CRF$_{UMN}$ | Changed F1 score | CancerBERT$_{UMN\_397}$ | Changed F1 score |
|---|---|---|---|---|---|---|
| Hormone Receptor type | 0.295 (0.646) | -0.649 (-0.308) | 0.448 (0.559) | -0.495 (-0.395) | 0.634 (0.842) | **-0.341\* (-0.139\*)** |
| Hormone Receptor status | 0.061 (0.061) | -0.481 (-0.481) | 0.000 (0.000) | -0.876 (-0.876) | 0.564 (0.564) | **-0.362\* (-0.362\*)** |
| Tumor size | 0.434 (0.582) | -0.260 (-0.167) | 0.659 (0.832) | **-0.052\* (0.019\*)** | 0.742 (0.921) | -0.122 (0.007) |
| Tumor site | 0.314 (0.568) | **0.110\* (-0.018\*)** | 0.220 (0.615) | -0.252 (-0.143) | 0.267 (0.736) | -0.334 (-0.050) |
| Tumor grade | 0.596 (0.596) | -0.285 (-0.329) | 0.575 (0. 575) | -0.341 (-0.341) | 0.733 (0.733) | **-0.194\* (-0.210\*)** |
| Tumor laterality | 0.001 (0.001) | -0.934 (-0.934) | 0.000 (0.000) | -0.952 (-0.952) | 0.620 (0.620) | **-0.339\* (-0.339\*)** |
| Cancer stage | 0.450 (0.450) | -0.418 (-0.418) | 0.310 (0.310) | -0.581 (-0.581) | 0.863 (0.863) | **-0.086\* (-0.086\*)** |
| Histological type | 0.435 (0.794) | -0.502 (-0.170) | 0.296 (0.718) | -0.638 (-0.241) | 0.552 (0.818) | **-0.382\* (-0.147\*)** |
| F1 Macro average | 0.323 (0.462) | -0.440 (-0.355) | 0.313 (0.451) | -0.524 (-0.438) | 0.622 (0.761) | **-0.270\* (-0.169\*)** |
| F1 Micro average | 0.354 (0.558) | -0.529 (-0.347) | 0.210 (0.328) | -0.683 (-0.594) | 0.616 (0.722) | **-0.309\* (-0.225\*)** |

Note: Texts in bold indicate the smallest change of F1 scores. The results were averaged scores based on 10 runs, * indicates the change of F1 score is statistically lower than other methods (CI: 0.95).

**Figure 2** shows the models' performances on different test sets. The original test set is the UMN test set, while portability test set is the MC test set. All models were UMN models (CRF$_{UMN}$, BiLSTM-CRF$_{UMN}$ and CancerBERT$_{UMN\_397}$) that trained solely on UMN data. The CancerBERT$_{UMN\_397}$ model got the most consistent performances across different test sets.

**Figure 2**. CRF$_{UMN}$, BiLSTM-CRF$_{UMN}$ and CancerBERT$_{UMN\_397}$ models' performances on different test sets. The original test set is the UMN test set, portability test set is the MC test set. All models were UMN models trained solely on UMN data.

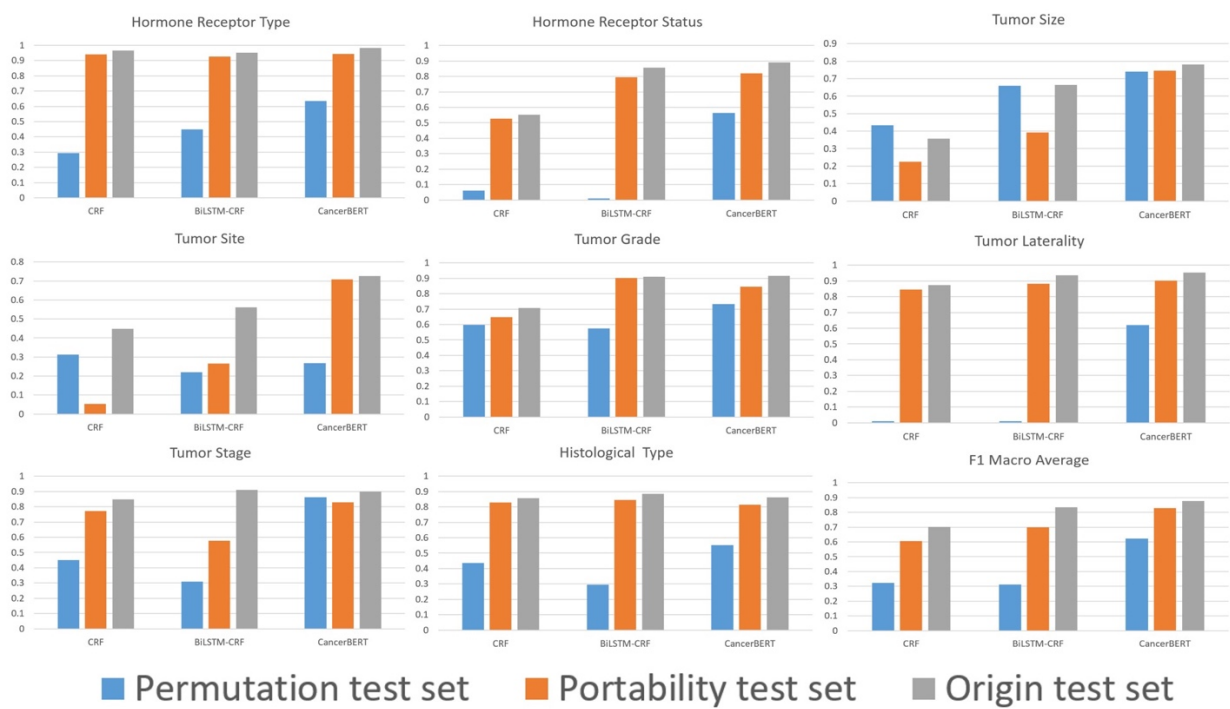

The performances of different types of models for entities under different ECR groups: 1) 0<= ECR <0.33, 2) 0.33<= ECR <0.67, 3) 0.67 <= ECR<1 and 4) ECR = 1 are shown in **Figure 3**. All three types of models achieved relatively good performances for groups 3&4. The CancerBERT$_{UMN\_397}$ model obtained significantly better performances in groups 1 & 2 compared to the other two models.

**Figure 3**. The performances of CRF$_{UMN}$, BiLSTM-CRF$_{UMN}$ and CancerBERT$_{UMN\_397}$ models for entities in different ECR groups. Group 1: 0<= ECR <0.33, Group 2: 0.33<= ECR <0.67, Group 3: 0.67 <= ECR<1 and Group 4: ECR = 1.

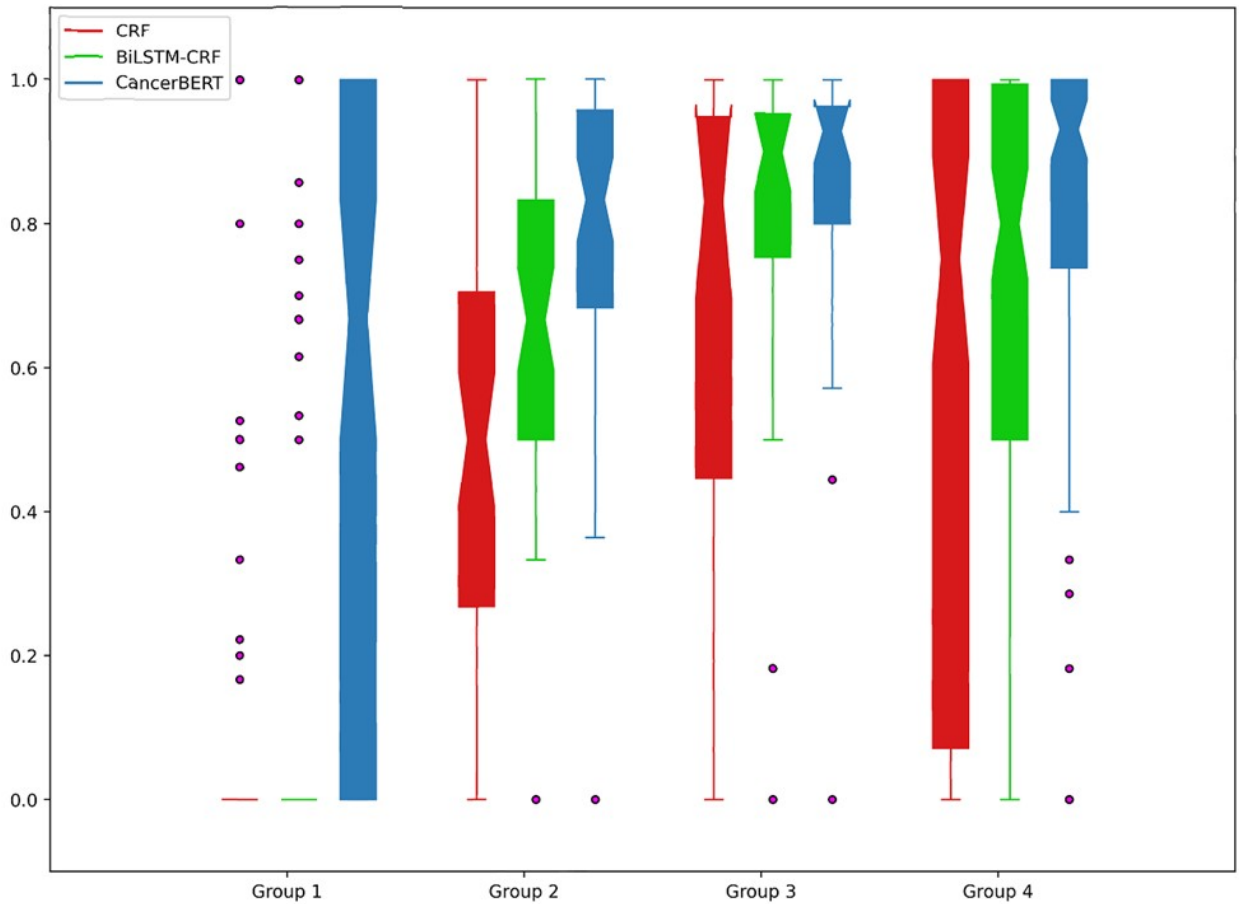

## Discussion

The clinical texts in EHR are different from general language and contain lots of professional phrases, medical jargons, acronyms and abbreviations. The semantic and contextual information of clinical texts in EHRs from different institutes usually vary a lot, and they should be seriously considered during the NLP algorithm development process. The corpora of UMN and MC were compared from various perspectives. More unique phenotypes were identified in UMN's annotation data (673 vs 592). The MC corpus has a higher density for the breast cancer phenotypes in the clinical texts (3.7 phenotypes/100 tokens vs 2.1 phenotypes/100 tokens). The similarities among breast cancer phenotypes are significantly higher than the similarity between the two corpora (0.9088 vs 0.5411), a similar phenomenon found in a previous study [1]. It indicates that clinicians may use similar clinical language to describe specific medical concepts in clinical texts, which lays the foundation to transport the NLP models among different clinical institutes.

In this study, we mainly explored the generalizability of the BERT-based models (CancerBERT models), along with other two classic machine learning models, i.e., CRF and BiLSTM-CRF. One important indicator of the model's generalizability is the portability of the model. We evaluated the model portability through the cancer phenotype extraction task. The results show that when directly evaluating the UMN models on the MC test data, only the CancerBERT models achieved reasonable performances, indicating their advantages in portability compared to the BiLSTM-CRF and CRF models. They also have the best overall performances after the models were refined based on the MC data, and they are more stable, the average drop of micro F1 score (CancerBERT$_{MC\_refined}$ vs CancerBERT$_{UMN}$) is 0.067 for CancerBERT models, for CRF and BiLSTM-CRF, the drop of scores are 0.078 (CRF$_{MC\_refined}$ vs CRF$_{UMN}$) and 0.092 (BiLSTM-CRF$_{MC\_refined}$ vs BiLSTM-CRF$_{UMN}$). Although the CancerBERT$_{MC\_397}$ model trained from scratch using MC data obtained the best overall performances, the performances of the CancerBERT$_{MC\_refined}$ model is comparable (F1 score is 0.007 lower than) to the CancerBERT$_{MC\_397}$ model. It indicates that BERT-based models can be transported to another clinical institute with keeping relatively high level of performance using a very low cost, since only a few annotated data are needed to fine-tune the model instead of training from scratch with a larger amount of corpus. From practical perspective, the CancerBERT$_{MC\_397}$ model could identify 37.1 phenotypes per patient in our dataset, while the CancerBERT$_{MC\_refined}$ model identified 36.6 (0.44 less) phenotypes per patient. As a comparison, the best CRF$_{MC\_refined}$ and BiLSTM-CRF$_{MC\_refined}$ model identified 31.1 and 34.0 phenotypes per patient which are 6 and 3.1 less phenotypes per patient. We calculated the correlation between the similarities of cancer phenotypes of two corpora and the performance drops (UMN models evaluated on the UMN test set and UMN models directly evaluated on the MC test set). The Pearson correlation scores are -0.678, -0.345 and -0.712 for CRF$_{UMN}$, BiLSTM-CRF$_{UMN}$ and CancerBERT$_{UMN\_397}$ models, respectively, indicating from medium (BiLSTM-CRF$_{UMN}$) to strong (CRF$_{UMN}$, CancerBERT$_{UMN\_397}$) negative correlations. The results indicate the portability of deep learning models is associated with the similarity among the targeted entities (e.g., cancer phenotypes in this study) from different corpora.

The evaluation on the permutation test set reflects whether the model is learning the real patterns in the texts or just memorizing the phenotypes that appear in the training set. In this permutation set evaluation, the CancerBERT model also shows a huge advantage compared to the CRF and BiLSTM-CRF models. For example, **Table 4** shows that the exact match (lenient match) macro-average F1 score only dropped 0.270 (0.169) for the CancerBERT$_{UMN\_397}$ model, while for CRF$_{UMN}$ and BiLSTM-CRF$_{UMN}$ models, the corresponding macro-average F1 scores drop by 0.440 (0.355) and 0.524 (0.438), respectively. Practically, these results show that among the three types of models, the CancerBERT model with the most robustness could better learn the contextual information of the entities, thus it could better handle new entity variants that were unseen before.

The ECR was applied to analyze the portability evaluation results. Though the CancerBERT$_{UMN\_397}$ model has significantly better performances in all groups, it has the biggest advantages for phenotypes in Groups 0 and 1 (ECR<0.67). These groups contain the target test phenotypes that are either appearing in the training sets with different labels, or are not in the training sets, which indicates the phenotypes in the two groups are relatively difficult to extract. The results indicate the CancerBERT$_{UMN\_397}$ model has a better learning ability to learn the target phenotypes and their contexts compared to the BiLSTM-CRF$_{UMN}$ and CRF$_{UMN}$ models.

The study has several limitations. We evaluated the generalizability of machine learning models through an NLP task to extract the breast cancer phenotypes from clinical texts among two health institutes. It is a common NER task in the clinical domain; however, there are many other NLP tasks, for example, relation extraction and text classification. It is worth for additional investigation on other clinical NLP tasks. We were focusing on the performances of the NLP models for the NER task, and how differences of NLP performances will further influence the downstream clinical applications remains to be explored in the future. We evaluated the portability only on two clinical corpora, and further exploration involving corpus from more institutes would help us better understand the NLP model generalizability.

## CONCLUSIONS

In this study, we evaluated the generalizability of three types of NLP models (CRF, BiLSTM-CRF and CancerBERT) in the clinical domain through an information extraction task to extract breast cancer phenotypes in clinical texts obtained from the UMN and the MC. The models' generalizability was evaluated from different perspectives, and the CancerBERT models were found to have the best generalizability since they could better learn the contextual information of target phenotypes, and deal with the textual variations in clinical texts. Our results show the CancerBERT models trained on a single institute can be transferred to other institutes and achieved comparable performance at low costs. Also, the CancerBERT models were found to have the greatest robustness among the three models. This study is the first study to analyze the portability of BERT-based models in the clinical domain and our findings are meaningful to guide the transfer and adoption of NLP models among different clinical institutes.

## ACKNOWLEDGEMENTS

NA

## DATA AVAILABILITY

The data underlying this article cannot be shared publicly due to the privacy of patient health information.

## FUNDING STATEMENT

This work is partially supported by the National Center for Complementary and Integrative Health (NCCIH) under grant number R01AT009457 (Zhang), National Institute on Aging under grant number 1R01AG078154-01 (Zhang); and the University of Minnesota Clinical and Translational Science Institute (CTSI), supported by the National Center for Advancing Translational Sciences under grant number UL1TR002494. The content is solely the responsibility of the authors and does not necessarily represent the official views of the NIH.

## COMPETING INTERESTS STATEMENT

There are no competing interests.

## CONTRIBUTIONSHIP STATEMENT

SZ conducted the main experiments. NW, SZ and LW participated in the data collection and annotation. HL and RZ guided the study design, data collection and analysis. All authors participated in the writing of the manuscript and critical revisions of the manuscript for important intellectual content.

**Abbreviations:**

EHR: electronic health records

NLP: natural language processing

UMN: University of Minnesota

BERT: bidirectional encoder representations from transformers

MC: Mayo Clinic

CRF: conditional random field

BiLSTM-CRF: bidirectional long short-term memory CRF

IRB: Institutional Review Board

ECR: entity coverage ratio

IAA: inter-annotator agreement

TF-IDF: term frequency inverse document frequency

NER: name entity extraction

GloVe: Global Vectors for Word Representation